\documentclass{new_tlp}

\usepackage[T1]{fontenc}
\usepackage{mathptmx}
\usepackage{amssymb}
\usepackage{amsmath}
\usepackage{color}
\usepackage[ruled,vlined,linesnumbered]{algorithm2e}
\usepackage[draft]{fixme}
\usepackage{tikz}
\usepackage[caption=false]{subfig}
\usepackage{pgfplots}
\usepackage{booktabs}
\usepackage{paralist}
\usepackage{xifthen}
\usepackage{asplisting}

\usepackage{url}

\usepackage{listings}

\fxsetup{
	inline,
	nomargin,
	theme=color,
	mode=multiuser
}
\FXRegisterAuthor{ks}{aks}{\color{blue}KS}
\FXRegisterAuthor{fr}{afr}{\color{green}FR}
\FXRegisterAuthor{cd}{acd}{\color{magenta}CD}
\FXRegisterAuthor{pg}{apg}{\color{orange}PG}
\FXRegisterAuthor{bm}{abm}{\color{red}BM}

\SetFuncSty{textrm}
\SetKwFunction{GetCore}{ComputeUnsatCore}
\SetKwFunction{MinimizeCore}{MinimizeCore}
\SetKwFunction{GetQuery}{DetermineQuery}
\SetKwFunction{GetDiagnoses}{ComputeDiagnoses}
\SetKwInput{Input}{input}

\pgfplotsset{
	filter discard warning=false 
	, legend cell align=left
	, minor grid style={loosely dotted, lightgray}
	, major grid style={loosely dashed, lightgray}
}
\pgfplotscreateplotcyclelist{cactus}{%
	red,      mark size=2.0pt, mark=star\\%
	blue,        mark size=2.0pt, mark=*\\%
	darkgreen,       mark size=2.0pt, mark=diamond*\\%
	black!80, mark size=2.0pt, mark=triangle*\\%
	gray,    mark size=2.0pt, mark=square\\%
}

\newcommand{\tabref}[1]{Table~\ref{#1}}

\newcommand{\myParagraph}[1]{\paragraph{#1}}

\def\naf{\ensuremath{\raise.17ex\hbox{\ensuremath{\scriptstyle\mathtt{\sim}}}}\xspace}
\newcommand{\hclasp}{\textsc{hclasp}\xspace}
\newcommand{\clasp}{\textsc{clasp}\xspace}
\newcommand{\measp}{\textsc{me-asp}\xspace}
\newcommand{\claspfolio}{\textsc{claspfolio}\xspace}
\newcommand{\wasp}{\textsc{wasp}\xspace}
\newcommand{\gringo}{\textsc{gringo}\xspace}
\newcommand{\minisat}{\textsc{minisat}\xspace}

\newcommand{\vbs}{\textsc{vbs}\xspace}

\newcommand{\inco}{\textsc{inconsistent}\xspace}

\newcommand{\event}[2]{\textsc{On#1\ifthenelse{\isempty{#2}}{}{(#2)}}\xspace}
\newcommand{\response}[2]{\textsc{\##1(#2)}\xspace}

\newcommand{\pup}{PUP\xspace}
\newcommand{\ucap}{\textsc{UCap}\xspace}
\newcommand{\iucap}{\textsc{IUCap}\xspace}
\newcommand{\qpup}{\textsc{Quick\pup}\xspace}
\newcommand{\qspup}{$\textsc{Quick\pup}^*$\xspace}
\newcommand{\pred}{\textsc{Pred}\xspace}

\newcommand{\ccp}{CCP\xspace}
\newcommand{\accp}[1]{A#1\xspace}
\newcommand{\mccp}{A1A2\xspace}
\newcommand{\bccp}{A2F\xspace}
\newcommand{\bfccp}{A2FO\xspace}
\newcommand{\bfaccp}{A2AFO\xspace}
\newcommand{\enc}[1]{\textsc{Enc#1}\xspace}

\newcommand{\cpp}{\textsc{c++}\xspace}
\newcommand{\pyth}{\textsc{python}\xspace}
\newcommand{\perl}{\textsc{perl}\xspace}

\definecolor{darkgreen}{RGB}{68,180,46}
\definecolor{darkgray}{RGB}{80,80,80}

\def\naf{\ensuremath{\raise.17ex\hbox{\ensuremath{\scriptstyle\mathtt{\sim}}}}\xspace}
\def\A{\ensuremath{\mathcal{A}}\xspace}

\def\I{\ensuremath{\mathcal{I}}\xspace}
\def\F{\ensuremath{\mathcal{F}}\xspace}
\def\S{\ensuremath{\mathcal{S}}\xspace}

\title[Combining Answer Set Programming and Domain Heuristics]
{Combining Answer Set Programming and Domain Heuristics for Solving Hard Industrial Problems\\
(Application Paper)}

\author[C. Dodaro et al.]{Carmine Dodaro$^1$, Philip Gasteiger$^2$, Nicola Leone$^1$, \and Benjamin Musitsch$^2$, Francesco Ricca$^1$, Kostyantyn Shchekotykhin$^2$\\
$^1$Department of Mathematics and Computer Science, University of Calabria, Italy \\
\email{\{lastname\}@mat.unical.it} \\
$^2$Alpen-Adria-Universit\"at Klagenfurt, Austria  \email{\{firstname.lastname\}@gmail.com}
}

\jdate{March 2003}
\pubyear{2003}
\pagerange{\pageref{firstpage}--\pageref{lastpage}}
\doi{S1471068401001193}

\begin{document}
	\label{firstpage}
	
	\maketitle
	
\begin{abstract}
Answer Set Programming (ASP) is a popular logic programming paradigm that has been applied for solving a variety of complex problems.
Among the most challenging real-world applications of ASP are two industrial problems defined by Siemens: 
the Partner Units Problem (\pup) and the Combined Configuration Problem (\ccp).
The hardest instances of \pup and \ccp are out of reach for
state-of-the-art ASP solvers. 
Experiments show that the performance of ASP solvers could be significantly improved by embedding domain-specific heuristics, but a proper effective integration of such criteria in off-the-shelf ASP implementations is not obvious.
In this paper the combination of ASP and domain-specific heuristics is studied with the goal of effectively solving real-world problem instances of \pup and \ccp. 
As a byproduct of this activity, the ASP solver \wasp was extended with an interface that eases embedding new external heuristics in the solver.
The evaluation shows that our domain-heuristic-driven ASP solver finds solutions for all the real-world instances of \pup and \ccp ever provided by Siemens.
\end{abstract}


\section{Introduction}

Answer Set Programming (ASP)~\cite{DBLP:journals/cacm/BrewkaET11} is a programming paradigm which has been proposed in the area of logic programming and non-monotonic reasoning.
ASP has become a popular paradigm for solving complex problems since it combines high knowledge-modeling power~\cite{DBLP:journals/csur/DantsinEGV01} with robust solving technology.
ASP has been applied in the areas of Artificial Intelligence~\cite{DBLP:conf/lpnmr/BalducciniGWN01,DBLP:journals/tplp/GagglMRWW15,DBLP:conf/lpnmr/BaralU01,DBLP:journals/tplp/ErdemPSSU13}, 
Bioinformatics~\cite{DBLP:journals/tplp/ErdemO15,DBLP:journals/tplp/KoponenOJS15,DBLP:journals/jetai/CampeottoDP15}, 
and Databases~\cite{DBLP:journals/dke/MarileoB10,DBLP:journals/tplp/MannaRT15}, to mention a few.
ASP has also been attracting the interest of companies~\cite{DBLP:conf/birthday/GrassoLMR11,InvitedFriedrich15}.

Among the most challenging industrial applications of ASP is the solution of two problems defined by Siemens, namely the Partner Units Problem (\pup)~\cite{DBLP:journals/aiedam/FalknerHSS08} and the Combined Configuration Problem (\ccp)~\cite{DBLP:conf/lpnmr/GebserRS15}.
The \pup originates from the railway safety domain and it generalizes a variety of other industrial problems in the area of security monitoring, peer-to-peer networking, etc.~\cite{DBLP:conf/iaai/TeppanFF12}.
The \ccp abstracts a family of configuration problems recurring in the practice of Siemens with application in systems of railway interlocking, safety automation, resource distribution, etc.
Given their practical importance, \pup and \ccp have been the subject of research in the last few years~\cite{DBLP:journals/aiedam/FalknerHSS08,DBLP:conf/ijcai/AschingerDGJT11,DBLP:conf/iaai/TeppanFF12,DBLP:conf/ictai/Drescher12,DBLP:conf/lpnmr/GebserRS15}.

The solution of these problems using ASP was quite successful from the perspective of modeling.
Indeed, both natural and optimized logic programs for \pup and \ccp were proposed in~\cite{DBLP:conf/cpaior/AschingerDFGJRT11,DBLP:conf/lpnmr/GebserRS15,DBLP:journals/ai/CalimeriGMR16}.
However, state-of-the-art ASP solvers (including portfolios) fail to compute solutions of the hardest real-world instances provided by Siemens in a reasonable amount of time (where finding a solution is the main real-world requirement).
Indeed, the instances of \pup and \ccp were among the hardest benchmarks of the recent ASP Competitions \cite{aspweb,DBLP:journals/ai/CalimeriGMR16}. 
One possible approach for boosting the performance of an ASP solver on specific problems is to provide a domain-specific heuristic~\cite{DBLP:conf/aaai/GebserKROSW13}.
This idea can be particularly effective for solving real-world industrial problems. As stated by Gerhard Friedrich in his joint invited talk at CP-ICLP 2015 ``domain-specific heuristics turned out to be  the key component in several industrial applications of problem solvers''~\cite{InvitedFriedrich15}.
The first attempts of solving \ccp instances by combining domain-heuristics and an off-the-shelf ASP implementation have been very promising~\cite{DBLP:conf/lpnmr/GebserRS15}. Nonetheless, the challenge of solving the hardest real-world instances using ASP remained open~\cite{DBLP:conf/lpnmr/GebserRS15}.

In this paper, the combination of ASP and domain-specific heuristics is studied with the goal of providing effective solutions to \pup and \ccp. 
Several heuristic criteria are considered, including the ones already proposed, as well as novel ones. 
The heuristics were implemented by extending the ASP solver \wasp~\cite{DBLP:conf/lpnmr/AlvianoDLR15}, 
and their performance is empirically evaluated on the real-world instances of \pup and \ccp provided by Siemens.

One of the lesson learned dealing with the solution of \pup and \ccp is that finding an effective combination of domain-heuristics with ASP is not obvious.
On the one hand, heuristic criteria are usually invented by domain experts studying the properties of solutions.
Thus, it is likely that such criteria do not fit the working principles of solvers and are not always embeddable in ASP encodings. 
On the other hand, despite the fact that an average developer could easily come up with an implementation of a heuristic criterion, the final embedding of an external algorithm in ASP solvers requires in depth knowledge of the internals of ASP implementations, which are nowadays very optimized and sophisticated.

Therefore, the paper provides a pragmatic contribution in this respect. Indeed, as a byproduct of the research on \pup and \ccp, \wasp was extended with an interface that makes it easier for developers to embed new external heuristics in the solver. 
The interface provides simplified access to the procedures that repeatedly needed modifications in our work.
The extension of \wasp presented in this paper offers a wide range of possibilities to end users for developing new heuristics by implementing the above-mentioned interface.
In particular, it offers multi-language support including scripting languages for fast prototyping, as well as an embedded \cpp interface for performance-oriented implementations.
In our experience the implementation of several heuristics resulted to be much easier (with respect to the development of several ad-hoc modification) after \wasp was extended with the new interface.

\medskip
\noindent The contributions of this paper can be summarized as follows:

\begin{enumerate}
\item The combination of ASP and domain heuristics for solving \pup and \ccp is studied and effective solutions are presented (see Section~\ref{sec:heur}).
\item An extension of \wasp with an interface that eases the design and implementation of new external heuristics is presented (see Section~\ref{sec:implementation}). 
\item An experiment is carried out to both study the behavior of \wasp equipped with several domain heuristics and compare it to state-of-the-art solutions (see Section~\ref{sec:experiments}).
\end{enumerate}

The results of the experiments are positive: A combination of ASP and domain heuristics can be used to solve effectively both \pup and \ccp. 
Notably, \wasp empowered with the best considered domain-heuristic was able to solve all the hardest real-world instances ever provided by Siemens including those for which no solution was known before.

\section{Preliminaries} \label{sec:prelim}
In this section, Answer Set Programming (ASP) language and contemporary solving techniques are over-viewed first; then the reader is provided with a description of the problems approached in this paper, namely the Partner Units (\pup) and the Combined Configuration Problem (\ccp). Hereafter the reader is assumed to be familiar with ASP~\cite{DBLP:journals/cacm/BrewkaET11}.

\subsection{Answer Set Programming}
\myParagraph{Syntax and Semantics.}
Let $\A$ be a fixed, countable set of propositional atoms including $\bot$. 
A literal $\ell$ is either an atom $a$ (a positive literal), or an atom preceded by the negation as failure symbol $\naf$ (a negative literal).
The atom associated to $\ell$ is denoted by $atom(\ell)=a$.
The complement of $\ell$ is denoted by $\overline{\ell}$, where $\overline{a} = \naf a$ and $\overline{\naf a} = a$ for an atom $a$.
For a set $L$ of literals, $\overline{L} := \{\overline{\ell} \mid \ell \in L\}$, $L^+ :=L \cap \A$, and $L^- := \overline{L} \cap \A$.
A program $\Pi$ is a finite set of rules. A rule is an implication $a \leftarrow l_1, \dots , l_n$,
where $a$ is an atom, and $l_1,\dots,l_n$ are literals, $n \geq 0$. 
For a rule $r$, $H(r) = \{a\}$ is called the head of $r$ and $B(r)=\{l_1,\dots,l_n\}$ is called the body of $r$.
A rule $r$ is a fact if $B(r) = \emptyset$, and is a constraint if $H(r)=\{\bot\}$. 
A (partial) interpretation is a set of literals $I$ containing $\naf \bot$. 
$I$ is inconsistent if $I^+ \cap I^- \neq \emptyset$, otherwise $I$ is consistent.
$I$ is total if $I^+ \cup I^- = \A$.
Given an interpretation $I$, a literal $\ell$ is true if $\ell \in I$; is false if $\overline{\ell} \in I$, and is undefined otherwise. 
An interpretation $I$ satisfies a rule $r$ if $I \cap (H(r) \cup \overline{B(r)}) \neq \emptyset$.
Let $\Pi$ be a program, a model $I$ of $\Pi$ is a consistent and total interpretation that satisfies all rules in $\Pi$.
The reduct of $\Pi$ w.r.t. $I$ is the program $\Pi^{I}$ obtained from $\Pi$ by (i) deleting all rules $r$ having $B(r)^- \cap \ I \neq \emptyset$, and (ii) deleting the negative body from the remaining rules~\cite{DBLP:journals/ngc/GelfondL91}.
A model $I$ of a program $\Pi$ is an answer set if there is no model $J$ of $\Pi^{I}$ such that $J^+ \subset I^+$.
A program $\Pi$ is coherent if it admits answer sets, otherwise it is incoherent.

\myParagraph{Answer Set Computation.}
\label{sec:cdcl}
The computation of answer sets can be carried out by employing an extended version of the Conflict-Driven Clause Learning (CDCL) algorithm, which was introduced for SAT solving~\cite{DBLP:series/faia/2009-185}.
The algorithm takes as input a program $\Pi$, and outputs an answer set if $\Pi$ is consistent, otherwise it outputs $\{\bot\}$.

The computation starts by applying polynomial simplifications to strengthen and/or remove redundant rules on the lines of methods employed by SAT solvers~\cite{DBLP:series/faia/2009-185}.
After the simplifications step, the partial interpretation $I$ is set to $\{\naf \bot\}$, and the backtracking search starts.
First, $I$ is extended with all the literals that can be deterministically inferred by applying some inference rule (propagation step).
In particular, in ASP three propagation rules are applied: unit, support and unfounfed-free.
Roughly, unit propagation is as in SAT whereas support propagation and unfounded-free propagation are ASP-specific~\cite{DBLP:conf/lpnmr/GebserKK0S15,DBLP:conf/lpnmr/AlvianoDLR15}.
Three cases are possible after a propagation step is completed:
$(i)$ $I$ is consistent but not total. In that case, an undefined literal $\ell$ (called branching literal) is chosen according to some heuristic criterion, and is added to $I$. Subsequently, a propagation step is performed that infers the consequences of this choice.
$(ii)$ $I$ is inconsistent, thus there is a conflict, and $I$ is analyzed. 
The reason of the conflict is modeled by a fresh constraint $r$ that is added to $\Pi$ (learning).
Moreover, the algorithm backtracks (i.e. choices and their consequences are undone) until the consistency of $I$ is restored.
The algorithm then propagates inferences starting from the fresh constraint $r$. 
Otherwise, if the consistency of $I$ cannot be restored, the algorithm terminates returning $\{\bot\}$.
Finally, in case $(iii)$ $I$ is consistent and total, the algorithm terminates returning $I$.

State-of-the-art CDCL implementations usually employ a heuristic for choosing the branching literal that can be seen as a variant of the \minisat~\cite{DBLP:conf/sat/EenS03} heuristic. 
The \minisat heuristic is based on the \textit{activity} value, that is initially set to 0 for each atom.
Whenever a literal $\ell$ occurs in a learned constraint, the activity of $atom(\ell)$ is incremented by a value $inc$.
Then, after each learning step, the value of $inc$ is multiplied by a constant slightly greater than 1, to promote variables that occur in recently-learned constraints.
Once a choice is needed, the literal $\naf a$ is chosen, where $a$ is the undefined atom having the highest activity value (ties are broken randomly).
In the following, the \minisat~\cite{DBLP:conf/sat/EenS03} heuristic is referred to as the default branching heuristic of a solver.

For the sake of completeness, we mention that the CDCL algorithm is usually complemented with heuristics that control the number of learned constraints, and restart the computation to explore different branches of the search tree~\cite{DBLP:series/faia/2009-185}.

\subsection{Partner Units Problem (PUP)}
The \pup originates from the railway safety domain~\cite{DBLP:journals/aiedam/FalknerHSS08} and has a variety of other applications, including security monitoring systems and peer-to-peer networks.~\cite{DBLP:conf/iaai/TeppanFF12}. 
To ensure the safety of the train traffic, railway tracks (see Figure~\ref{fig:pup_layout}) are equipped with hardware \emph{sensors}, e.g.\ $s_1,\dots,s_6$, registering wagons passing by.
These sensors are organized in safety \emph{zones}, e.g.\ $z_1,\dots,z_{24}$.
In order to prevent wagons of different trains from entering the same zone simultaneously, the safety system has a set of \emph{control units}, e.g. $u_1, u_2, u_3$, that enforce safety requirements on connected zones and sensors.
Consequently, every PUP instance consists of 
\begin{inparaenum}[(i)]
    \item a layout of sensors $S$ and zones $Z$ represented as an undirected bipartite graph $G=(S,Z,E)$, where $S \cup Z$ are vertices connected by edges $E \subseteq S \times Z$;
    \item a set of available units $U$;
    \item a natural number $\ucap$ equal to the maximum number of sensors/zones connected to a unit; and
    \item a natural number $\iucap$ equal to the maximum number of inter-unit connections.
\end{inparaenum}

A solution of a PUP instance is an assignment of zones and sensors to units and an interconnection of units, such that
\begin{inparaenum}[(i)]
	\item every unit is connected to at most \ucap sensors and at most \ucap zones;
	\item every unit is connected to at most \iucap \emph{partner units}; and
	\item if a sensor $s$ is part of a zone $z$, then $s$ must be connected to the same or a partner unit of the unit connected to $z$.
\end{inparaenum}

\subsection{Combined Configuration Problem (CCP)}

The \ccp abstracts a number of real-world problems including railway interlocking systems, safety automation, and resource distribution.
It occurs in practical applications at Siemens, where a complex problem is composed of a set of subproblems~\cite{DBLP:conf/lpnmr/GebserRS15}.

A \ccp instance (see Figure~\ref{fig:ccp}) consists of
\begin{inparaenum}[(i)]
  \item a directed acyclic graph $G = (V, E)$, where each vertex has a type with an associated size, e.g. $b$ and $p$ with sizes $1$ and $3$ resp.;
  \item two disjoint paths $P_1$ and $P_2$, e.g. edges highlighted in red and green;
  \item a set of safety areas and their border elements, e.g. $A_1 = \{ b_1, b_2, b_3, p_1 \}$ and $A_2 = \{ b_2, b_4, b_5, p_3 \}$ with border elements $BE_1 = \{ b_1, b_2, b_3 \}$ and $BE_2 = \{ b_2, b_4, b_5 \}$;
  \item the maximum number $M$ of assigned border elements per area, e.g. $M = 3$;
  \item the number $C$ of colors, e.g. $C = 3$;
  \item the number $B$ of bins available per color, e.g. $B = 2$; and
  \item the capacity $K$ of each bin, e.g. $K = 3$.
\end{inparaenum}
A solution is an assignment of colors to vertices, vertices to bins and border elements to areas, such that the following subproblems are solved:
\begin{itemize}[(P5)]
  \item[(P1)] \emph{Coloring:}
  Every vertex must have exactly one color.
	
  \item[(P2)] \emph{Bin-Packing:}
  For each set $V_i \subseteq V$ of color $i$, assign each $v \in V_i$ to exactly one bin s.t.  for each bin the sum of vertex sizes is not greater than $K$, and at most $B$ bins are used.
	
  \item[(P3)] \emph{Disjoint Paths:}
  The color of any two vertices $u \in P_1$ and $v \in P_2$ must be different.
	
  \item[(P4)] \emph{Matching:}
  For each area $A$ assign a set of border elements $BE$, such that all border elements have the same color and $A$ has at most $M$ border elements. Additionally, each border element must be assigned to exactly one area.
  	
  \item[(P5)] \emph{Connectedness:}
  Two vertices of the same color must be connected via a path that comprises only vertices of this color.
\end{itemize}

\begin{figure}[t!]
\figrule
\begin{minipage}{.48\textwidth}
    \centering
       \begin{tikzpicture}[x=2em,y=2em,baseline=0pt]
       \tikzstyle{track} =   [double=black!10, double distance=0.2em, rounded corners=0.4em]
       \tikzstyle{sensor} =  [circle, fill, inner sep=0, minimum size=0.25em]
       \tikzstyle{zone} =    [black!60, dashed, semithick, rounded corners=0.2em]
       
       \draw[track] (0,0) -- ++(9,0)
       (0,0) -- ++(1.5,0) -- ++(1.5,1) -- ++(3,0) -- ++(1.5,-1) -- ++(1.5, 0);
       
       \node[sensor, label=below:$s_1$] at (1,-0.25) {};
       \node[sensor, label=above:$s_2$] at (2.25, 0.75) {};
       \node[sensor, label=below:$s_3$] at (2.25,-0.25) {};
       \node[sensor, label=above:$s_4$] at (6.75, 0.75) {};
       \node[sensor, label=below:$s_5$] at (6.75,-0.25) {};
       \node[sensor, label=below:$s_6$] at (8,-0.25) {};
       
       \draw[zone] (0.25,-0.85) rectangle node[left, label=left:$z_{1}$]{} ++(1,1);
       \draw[zone] (0.7,-1) rectangle node[below=1.5em, label=below:$z_{123}$]{} ++(1.8,2.45);
       \draw[zone] (1.7,0.6) rectangle node[above, label=above:$z_{24}$]{} ++(5.6,0.75);
       \draw[zone] (1.7,-0.85) rectangle node[below, label=below:$z_{35}$]{} ++(5.6,1);
       \draw[zone] (6.45,-1) rectangle node[below=1.5em, label=below:$z_{456}$]{} ++(1.8,2.45);
       \draw[zone] (7.75,-0.85) rectangle node[right, label=right:$z_{6}$]{} ++(1,1);
       
       \end{tikzpicture}
       
       \vspace{5pt}
       
       \begin{tikzpicture}[x=1.7em,y=1.7em,baseline=0pt]
       \tikzstyle{sensor} = [draw, circle, inner sep=0em, minimum width=1.5em]
       \tikzstyle{zone} =   [draw, rectangle, inner sep=.2em, minimum width=2em, rounded corners=0.2em]
       \tikzstyle{unit} =   [draw, rectangle, inner sep=.2em, minimum width=2em]
       
       \node[sensor] (s1) at (0.0,1.25) {$s_1$};
       \node[sensor] (s3) at (1.5,1.25) {$s_3$};
       \node[sensor] (s5) at (3.0,1.25) {$s_5$};
       \node[sensor] (s4) at (4.5,1.25) {$s_4$};
       \node[sensor] (s6) at (6.0,1.25) {$s_6$};
       \node[sensor] (s2) at (7.5,1.25) {$s_2$};

       \node[zone] (z1)   at (0.0,-1.25) {$z_{1}$};
       \node[zone] (z35)  at (1.5,-1.25) {$z_{35}$};
       \node[zone] (z123) at (3.0,-1.25) {$z_{123}$};
       \node[zone] (z456) at (4.5,-1.25) {$z_{456}$};
       \node[zone] (z6)   at (6.0,-1.25) {$z_{6}$};
       \node[zone] (z24)  at (7.5,-1.25) {$z_{24}$};
       
       \node[unit] (u1) at (1.50, 0) {$u_1$};
       \node[unit] (u2) at (3.75, 0) {$u_2$};
       \node[unit] (u3) at (6.00, 0) {$u_3$};

       \draw (z1)   -- (u1) -- (s1)
       (z35)  -- (u1) -- (s3)
       (z456) -- (u2) -- (s4)
       (z123) -- (u2) -- (s5)
       (z6)   -- (u3) -- (s6)
       (z24)  -- (u3) -- (s2);
       
       \draw (u1) -- (u2) -- (u3);
       \end{tikzpicture}
        \caption{Sample PUP instance (top) and a solution with $\ucap =\iucap=2$ (bottom).}               \label{fig:pup_layout}
\end{minipage}
\hfill
\begin{minipage}{.48\textwidth}
        \centering
        \tikzstyle{node} = [circle, draw, text centered, inner sep=0em, minimum width=15pt ]
        \tikzstyle{rnode} = [node, fill=red!20]
        \tikzstyle{bnode} = [node, fill=blue!20]
        \tikzstyle{gnode} = [node, fill=green!20]
        \tikzstyle{edge}  = [line width=1.3pt, ->]
        \tikzstyle{path1} = [edge,red]
        \tikzstyle{path2} = [edge,darkgreen]
        \tikzstyle{a1}    = [dashed]
        \tikzstyle{a2}    = [a1,draw=gray,very thick]
        
        \tikzset{
        	bin/.style={
        		label={[label distance=-4pt]-(-60):\scriptsize#1}}
        }
        
        \begin{tikzpicture}[auto]
        \node at (-1.6,0)    [node](b1) {$b_{1}$};    
        \node at (-0.8,0)    [node](p1) {$p_{1}$};    
        \node at (0,0)     [node](b2) {$b_{2}$};
        \node at (.8,0)     [node](p3) {$p_{3}$};
        \node at (1.6,0)     [node](b5) {$b_{5}$};
        
        \node at (-.8,-.9)    [node](b3) {$b_{3}$};    
        \node at (0,-.9)     [node](e1) {$p_{2}$};
        \node at (.8,-.9)     [node](b4) {$b_{4}$};
        
        \path[path1] (b1)  edge  node {} (p1);
        \path[edge]  (p1)  edge  node {} (b2);
        \path[edge]  (b2)  edge  node {} (p3);
        \path[path2] (p3)  edge  node {} (b5);
        \path[path1] (p1)  edge  node {} (b3);
        \path[edge]  (b3)  edge  node {} (e1);
        \path[edge]  (e1)  edge  node {} (b4);
        \path[path2] (b4)  edge  node {} (p3);
        
        \draw [rounded corners=10pt,dashed,darkgray,thick] (-1.9,0.5)--(-2.1,-0.2) -- node[below,fill=white] {\textcolor{black}{$\mathit{A}_1$}}  (-1.2,-1.3) -- (-.45,-1.3) -- (-0.5,-0.4) -- (0.35,-0.45) -- (.35,0.45)--cycle ;
        
        \draw [rounded corners=10pt,dashed,gray,very thick] (1.9,0.6)--(2.1,-0.2) -- node[below,fill=white] {\textcolor{black}{$\mathit{A}_2$}}  (1.2,-1.3) -- (.35,-1.3) -- (0.35,-0.45) -- (-0.45,-0.6) -- (-.45,0.6)--cycle ;
        
        \end{tikzpicture}
        
        \vspace{5pt}
        
        \begin{tikzpicture}[auto]
        \node at (-1.6,0)  [rnode,a1,bin={2}]   (b1) {$b_{1}$};    
        \node at (-0.8,0)  [rnode,bin={1}]      (p1) {$p_{1}$};    
        \node at (0,0)     [rnode,a1,bin={2}]   (b2) {$b_{2}$};
        \node at (.8,0)    [gnode,bin={1}]      (p3) {$p_{3}$};
        \node at (1.6,0)   [gnode,a2,bin={2}]   (b5) {$b_{5}$};
        
        \node at (-.8,-.9) [rnode,a1,bin={2}]   (b3) {$b_{3}$};    
        \node at (0,-.9)   [bnode,bin={1}]      (e1) {$p_{2}$};
        \node at (.8,-.9)  [gnode,a2,bin={2}]   (b4) {$b_{4}$};
        
        \path[path1] (b1)  edge  node {} (p1);
        \path[edge]  (p1)  edge  node {} (b2);
        \path[edge]  (b2)  edge  node {} (p3);
        \path[path2] (p3)  edge  node {} (b5);
        \path[path1] (p1)  edge  node {} (b3);
        \path[edge]  (b3)  edge  node {} (e1);
        \path[edge]  (e1)  edge  node {} (b4);
        \path[path2] (b4)  edge  node {} (p3);
        
        \end{tikzpicture}
        \caption{Sample \ccp instance (top) and its solution (bottom).}
        \label{fig:ccp}    
\end{minipage}
\figrule 
\end{figure}

\begin{table}[t!]
\footnotesize	
\begin{tabular}{lp{.7\textwidth}}
\toprule
  \multicolumn{2}{c}{\textbf{CDCL $\longrightarrow$  Heuristic: Events} } \\
\midrule
  \event{Search}{$\Pi',\A'$} &
  \textit{Triggered when the backtracking search starts. $\Pi'$ and $\A'$ are the program and the atoms after the simplifications, respectively.}\\
\addlinespace
	
  \event{IncoChoice}{$\ell$} &
  \textit{Triggered when the choice $\ell$ led to an inconsistency.}\\
\addlinespace
	
  \event{Conflict}{$\ell$} &
  \textit{Triggered when a conflict is detected. After analyzing the conflict the consistency is restored, i.e. choices and their consequences are undone. $\ell$ is the latest valid branching literal after the analysis.}\\
\addlinespace
	
  \event{LearnConstraint}{$c$} &
  \textit{Triggered when the constraint $c$ is learned.} \\
\addlinespace 
	
\event{LitTrue}{$\ell$} & \textit{Triggered when a literal $\ell$ is set to true.}\\
\addlinespace 
	
  \event{Restart}{~} &
  \textit{Triggered when a restart of the search occurs.}\\
\addlinespace 
	
  \event{UnrollLit}{$\ell$} &
  \textit{Triggered during backtracking if the literal $\ell$ is set to undefined.}\\
  
  \addlinespace
	\event{ChoiceRequired}{~} &
    \textit{Triggered when a choice is required.}\\

\midrule
\multicolumn{2}{c}{\textbf{Heuristics $\longrightarrow$  CDCL: Commands}}\\
\midrule
	\response{Choose}{$\ell$} & \textit{Choose $\ell$ as next branching literal.}\\
	\addlinespace
	\response{Unroll}{$\ell$} & \textit{Unroll the previous choices until $\ell$ becomes undefined. To trigger a restart set $\ell=\bot$.}\\
	\addlinespace
	\response{Fallback}{$n, \I, \F, \S$} & \textit{Use the default heuristics for the next $n$ choices.
			The default heuristic is enable permanently if $n \leq 0$.
			Parameters $\I, \F, \S$ initialize the default heuristics as follows:
			$\I: A \rightarrow \mathbb{N}$ provides the initial activity of atoms;
			$\F: A \rightarrow \mathbb{N}$ associates to atoms an amplifying factor;
			$\S: A \rightarrow \{pos,neg\}$ to provide a priority on the sign of literals.}\\ 
	\addlinespace
	\response{AddConstraint}{$c$} & \textit{Add the constraint $c$ to the program.}\\
	\bottomrule
\end{tabular}
\caption{Events of the solver and commands issued by the heuristic.}
\label{tab:eventscommands}
\end{table}

\section{Domain Heuristics}
\label{sec:heur}
In this section, we study several domain-heuristics for solving \pup and \ccp and their embedding in an ASP solver.
The implementation of different domain-heuristics required to extend the CDCL computation more or less the same in some specific points. 
The interface of methods reported in \tabref{tab:eventscommands} (and described below) abstracts some of these points and allows us to factor out the details of the heuristics from the main CDCL algorithm. 

\myParagraph{Heuristics Interface.}
The implementation of a heuristic may require information regarding the current state of the search.
To this end, \textit{events} (see \tabref{tab:eventscommands} upper part) are triggered when certain points of the CDCL algorithm are reached.
The status of the heuristic can be thus initialized or updated by handling events. 
For example, the \event{Search}{} event may be used to initialize some heuristic counters at the beginning of the search.
The \event{LitTrue}{} and \event{UnrollLit}{} events can be used to keep track of the current partial interpretation.
Whereas \event{IncoChoice}{}, \event{Conflict}{}, and \event{Restart}{} events might be used to handle conflicts and restarts, respectively.
The \event{LearnConstraint}{} event could be used to keep track of changes to the logic program that occur during the learning process.
To drive the search for a solution, the heuristic issues commands (see \tabref{tab:eventscommands} lower part). 
This is done whenever a new branching literal is required by the CDCL algorithm, i.e. whenever the \event{ChoiceRequired}{} event is triggered.
In more detail, the command \response{Choose}{$\ell$} instructs the solver to use $\ell$ as the next branching literal.
Since domain-specific information may allow to recognize a dead end in advance, the heuristics may add a new constraint $c$ (issuing \response{AddConstraint}{$c$}) or backtrack to an earlier choice $\ell$ (issuing \response{Unroll}{$\ell$}).
Finally, when the heuristic is unable to compute further choices, the \response{Fallback}{$n,\I,\F,\S$} command instructs the solver to use the default heuristic for $n$ steps.

Note that one may choose a different interface to implement heuristics in CDCL solvers.
The ones reported above are expressive enough to implement the domain heuristics for \pup and \ccp as detailed below, and to provide a high-level description of our implementations.

\subsection{PUP Heuristics}
\label{sec:pupheur}

A number of PUP heuristics are suggested in the literature~\cite{DBLP:conf/ijcai/AschingerDGJT11,Ryabokon2015}. 
We consider the \qpup~\cite{DBLP:conf/iaai/TeppanFF12} heuristic and its derivatives, since ad hoc algorithms based on them outperform the best off-the-shelf ASP solvers:

\begin{itemize}
\item
  \qpup first generates an order of zones and sensors.
  To do so, given the input graph $G = (S, Z, E)$, the heuristic traverses the edges in $E$ breadth-first, starting with some zone $z \in Z$.
  Each time a new branching literal is requested, the next unassigned zone or sensor with respect to the order is selected.
  Afterwards, the selected zone or sensor is assigned to a unit, preferring new units (i.e. units without any zones or sensors assigned).
  If no assignment is possible, the last one performed is unrolled and another assignment is tried.
  
\item 
  \qspup is a variation of \qpup that prefers used units instead of new units.

\item
  \pred~\cite{DBLP:conf/ictai/Drescher12} is a variation of \qpup that prefers units assigned to neighbors of the considered vertex (i.e. units with zones or sensors that are reachable within two edges in the input graph starting with the current zone or sensor).
\end{itemize}

The integration of the \pup heuristic in the CDCL algorithm is done using the interface presented in \tabref{tab:eventscommands}.
When \event{Search}{} is triggered, the heuristic is initialized.
This is done by parsing the atoms that represent sensors, zones, the zone-to-sensor relation, available units, \ucap, and \iucap.
Furthermore, the order of vertices is generated according to one of the heuristic criteria described above.
Whenever a branching literal is required, the order is used to determine the next unassigned zone or sensor.
This zone or sensor is then assigned to a unit by returning the \response{Choose}{$\ell$} command, where $\ell$ is the literal corresponding to the appropriate zone-to-unit or sensor-to-unit atom. 
The implementation keeps track of the already done assignments in a list.
If the solver retrieves a conflict, the \event{Conflict}{} event is triggered and the heuristic restores a consistent state.
In addition, \event{UnrollLit}{} and \event{LitTrue}{} are used to synchronize the internal state of the heuristics with the one of the CDCL algorithm.
Finally, the heuristic may recognize that the current partial assignment cannot be completed to a valid solution.
In this case, a restart is performed by returning the \response{Unroll}{$\bot$} command and subsequently adding a constraint $c$ preventing the partial assignment using the \response{AddConstraint}{$c$} command.

\subsection{CCP Heuristics}
\label{sec:ccpheur}
Various heuristics for computing partial solutions of the \ccp have been suggested in \cite{DBLP:conf/lpnmr/GebserRS15}.
However, in contrast with \pup, no complete heuristic for \ccp is known.
There are heuristics for the subproblems, e.g.\ best-fit for bin packing or minimal degree for graph  coloring~\cite{DBLP:books/fm/GareyJ79}, but the task of combining them is non-trivial.

The following heuristics are proposed in~\cite{DBLP:conf/lpnmr/GebserRS15}:
\begin{itemize}
\item 
	Algorithm 1 (\accp{1}) solves the matching subproblem (P4).
	It iterates over the set of border elements	and assigns them to the area with the lowest number of border elements so far.
\item
	Algorithm 2 (\accp{2}) solves the subproblems (P1), (P2) and (P5).
	It starts by creating a queue $Q$, initially filled with some arbitrary vertex $v \in V$ and setting the initial color $c$.
	Afterwards, the first vertex $v$ in $Q$ is colored with $c$ and placed into a bin according to the \emph{best-fit} heuristic.
	If successful, all uncolored neighbors of $v$ are added to $Q$; otherwise color of $v$ is removed.
	The process is repeated until the queue is empty.
	In this case, the algorithm gets the next color and adds some arbitrary uncolored vertex to $Q$.
\item
	\mccp applies both of the previous heuristics, extending the output of \accp{1} by using \accp{2}.
\end{itemize}

We elaborate in our study the following variants:
\begin{itemize}
\item
	\bccp is based on A2.
	A fallback to the default heuristic is done when
	\begin{inparaenum} [(i)]
	\item a partial solution is found, 
	\item the heuristic is used for 10 seconds, or
	\item the heuristic can not make further choices.
	\end{inparaenum}
	The latter case might occur if all atoms used in the subproblems have been assigned or ignored.
\item
	\bfccp extends \bccp with a descending ordering of all vertices in $G$ as well as in $Q$ w.r.t. a score.
	Vertices having either only incoming or only outgoing edges are rated with a score of 1, all other vertices are rated with a score of 0.
\item
	\bfaccp is a variant of \bfccp with an alternating fallback strategy between \bfccp and the solvers default heuristic.
	The heuristic strategy is switched every 10 seconds.
	When \bfccp is restarted, a new order is created.
	If all possible orders have been tried, \bfccp falls back to the default heuristic permanently.
	Intuitively, the default heuristic of the solver may become more effective when a number of representative conflicts are acquired.
\end{itemize}
The heuristics given above are implemented similarly to the ones for the \pup. 
First, when the \event{Search}{} event is triggered, (i) the input graph is read, and (ii) an empty queue $Q$ is created, which is used to extract the next choices.
In addition, the \bfccp and \bfaccp heuristics assign a score to the vertices of the input graph, as described above (all other heuristics assign a score of 0 to all vertices).
The set of all vertices (order) is sorted with respect to their score.
Whenever a branching literal is requested, a vertex $v$ is selected, where $v$ is the first element of the queue $Q$.
If the queue is empty, $v$ is the first uncolored and unassigned vertex among all vertices.
At this point, the heuristic computes a color or bin assignment for $v$, prioritizing the first one.
The atom $\ell$ corresponding to the obtained assignment is returned to the CDCL algorithm using the \response{Choose}{$\ell$} command.
If $v$ was successfully colored or assigned to a bin, (i.e. no conflict is implied by the choice), $Q$ is updated as follows: (i) $v$ is removed from $Q$, (ii) all non-processed neighbors of $v$ are added to $Q$, (iii) vertices in $Q$ are ordered according to their score.
Otherwise, if the solver finds a conflict (event \event{Conflict}{}), the heuristic restores a consistent state of the queue, order, and choices.
The color and bin assignments made by the CDCL algorithm are tracked by the heuristics using the \event{UnrollLit}{} and \event{LitTrue}{} events.
When the queue is empty and all vertices in the order have been considered, the control is given back to CDCL by using the \response{Fallback}{$0,\emptyset,\emptyset,\emptyset$} command to complete the partial assignment.
Finally, in the variants requiring alternation of heuristics with the default strategy, the \response{Fallback}{$1,\emptyset,\emptyset,\emptyset$} and \response{Unroll}{$\bot$} commands are issued every 10 seconds.

\section{Implementation}\label{sec:implementation}
We implemented an infrastructure for easy specification and testing of heuristic strategies as an extension of the ASP solver \wasp~\cite{DBLP:conf/lpnmr/AlvianoDLR15}.
\wasp accepts as input a propositional program encoded in the numerical format produced by the grounder \gringo \cite{DBLP:conf/lpnmr/GebserKKS11} and uses a CDCL-like algorithm for computing answer sets (see Section~\ref{sec:cdcl}).
Our extended version of \wasp can additionally ask for guidelines from an external module implementing a new heuristic.
The communication with external heuristics follows a synchronous message passing protocol. 
The protocol is implemented (as customary in object-oriented languages) by means of method calls. 
Basically, an external heuristic implementation must be compliant with a specific interface.
The methods of the interface correspond to the events abstracted in Table~\ref{tab:eventscommands}.
Whenever a specific point of the computation is reached the corresponding event is triggered, i.e., a method of the heuristic is called.
Some of the methods of the interface are allowed to return values that are interpreted by \wasp as commands.
For instance, whenever the next choice must be provided from the external heuristic (i.e., the event 
\emph{onChoiceRequired} is triggered) a method is called on the heuristic that computes and returns the next branching literal.

Our implementation supports heuristic strategies implemented in (i) \perl and \pyth for obtaining fast prototypes and (ii) \cpp in case better performances are needed.
Note that \cpp implementations must be integrated in the \wasp binary at compile time, whereas \perl and \pyth can be specified by means of text files given as parameters for \wasp, thus scripting-based heuristic does not require changes and recompilation of \wasp.

The source code and the documentation are available on the branch \emph{plugins} at \url{https://github.com/alviano/wasp}.

\section{Experimental Analysis}\label{sec:experiments}
In this section, we present an experiment assessing the performance of the described heuristics on the industrial instances of PUP and CCP provided by Siemens. All instances, encodings and binaries can be downloaded from \url{http://yarrick13.github.io/hwasp/}. 

The evaluation was done on a system equipped with an i7-3030K CPU, 64GB RAM and Ubuntu 11.10.
The solvers were run for at most 900 seconds on instances grounded by \gringo \cite{DBLP:conf/lpnmr/GebserKKS11}. 
The grounding time is not reported, as we focus on solver performance only.
On average, the grounder required 1 and 5 seconds to ground PUP and CCP instances, respectively.
We compared our approach with \clasp~\cite[v.\ 3.1.3]{DBLP:conf/lpnmr/GebserKK0S15}, \claspfolio~\cite[v.\ 2.2.0]{DBLP:journals/tplp/HoosLS14} and \measp~\cite[v. 2015]{DBLP:journals/tplp/MarateaPR14}.
The solvers were executed using the following configurations:  
\begin{inparaenum}[(1)]
	\item standard \clasp with 1 thread;
	\item standard \clasp with 10 threads;
	\item manually configured \clasp portfolio with 10 different configured threads, e.g. configurations like handy or crafty, as well as two random decision configurations;
	\item \claspfolio with 10 threads; and
	\item \measp with the 2015 ASP Competition configuration.
\end{inparaenum}
For the sake of simplification, for each instance the best result obtained by one of the solvers mentioned above is given by the \emph{Virtually Best Solver} (\vbs).

The PUP instances are categorized in \emph{double}, \emph{double-variant}, \emph{triple} and \emph{grid}.
The \emph{grid} instances comprise parts of real-world railway systems, whereas all others are synthetic.
Nevertheless, the instances of the first three types represent interesting topologies occurring frequently in the practice of Siemens~\cite{DBLP:conf/cpaior/AschingerDFGJRT11}.
We used two different encodings \enc{1}~\cite{DBLP:conf/cpaior/AschingerDFGJRT11} (Sect.\ 3.1, p.\ 6) and \enc{2} (ASP Competition 2015). The first one is natural and can be written by a trained student or even automatically generated from UML diagrams by tools like OOASP~\cite{DBLP:conf/lpnmr/FalknerRSS15}.
\enc{2} is a complex encoding comprising symmetry breaking and ordering rules resulting in a significant performance improvement on some PUP instances.
Some deterministic choices made by these heuristics interfere with our approach.
Thus, we removed rule 9 (line 14) from \enc{2} only in executions of \wasp variants.

The CCP instances are categorized in \emph{easy}, \emph{moderate} and \emph{hard}.
We focused on  \emph{hard} instances since these have not been solved in previous work.
There are two types of hard instances:
\begin{inparaenum}[(i)]
	\item a set of 20 real-world instances by~\cite{DBLP:conf/lpnmr/GebserRS15} and
	\item 16 new instances with grid-like input graph structure (\emph{grids}).
\end{inparaenum} 
The encoding used for \ccp is the same of~\cite{DBLP:conf/lpnmr/GebserRS15}.

\begin{figure*}[t]
	\figrule
	\centering
	
	\subfloat[]{\label{fig:eval:pup:cactus}%
		\begin{tikzpicture}[scale=0.7]
		\pgfkeys{%
			/pgf/number format/set thousands separator = {}}
		\begin{axis}[
		scale only axis
		, font=\small
		, x label style = {at={(axis description cs:0.5,0.03)}}
		, y label style = {at={(axis description cs:0.05,0.5)}}
		, xlabel={Number of instances}
		, ylabel={Execution time (s)}
		, xmin=0, xmax=36
		, ymin=0, ymax=920
		, legend style={at={(0.28,0.98)},anchor=north, draw=none,fill=none}
		, legend columns=1
		, width=0.6\textwidth
		, height=0.5\textwidth
		, ytick={0,100,200,300,400,500,600,700,800,900}
		, xtick={0,6,12,18,24,30,36}
		, major tick length=2pt
		]
		
		\addplot [mark size=2.5pt, color=red, mark=diamond, dashed, mark options=solid]  table[col sep=semicolon, y index=16] {./cactus-pup.csv}; 
		\addlegendentry{\vbs (\enc{1})}
		
		\addplot [mark size=2.5pt, color=black, mark=triangle]  table[col sep=semicolon, y index=17] {./cactus-pup.csv}; 
		\addlegendentry{\wasp \qpup (\enc{1})}
		
		\addplot [mark size=2.5pt, color=black, mark=diamond]  table[col sep=semicolon, y index=18] {./cactus-pup.csv}; 
		\addlegendentry{\wasp \qspup (\enc{1})}
		
		\addplot [mark size=2.5pt, color=black, mark=o]  table[col sep=semicolon, y index=19] {./cactus-pup.csv};
		\addlegendentry{\wasp \pred (\enc{1})}
		
		\addplot [mark size=2.5pt, color=red, mark=diamond*, dashed, mark options=solid]  table[col sep=semicolon, y index=6] {./cactus-pup.csv};
		\addlegendentry{\vbs (\enc{2})}
		
		\addplot [mark size=2.5pt, color=blue, mark=triangle*]  table[col sep=semicolon, y index=7] {./cactus-pup.csv}; 
		\addlegendentry{\wasp \qpup (\enc{2})}
		
		\addplot [mark size=2.5pt, color=blue, mark=diamond*]  table[col sep=semicolon, y index=8] {./cactus-pup.csv}; 
		\addlegendentry{\wasp \qspup (\enc{2})}
		
		\addplot [mark size=2.5pt, color=blue, mark=*]  table[col sep=semicolon, y index=9] {./cactus-pup.csv}; 
		\addlegendentry{\wasp \pred (\enc{2})}
		\end{axis}
		\end{tikzpicture}
	}
	\subfloat[]{\label{fig:eval:ccp:cactus}%
		\begin{tikzpicture}[scale=0.7]
		\pgfkeys{%
			/pgf/number format/set thousands separator = {}}
		\begin{axis}[
		scale only axis
		, font=\small
		, x label style = {at={(axis description cs:0.5,0.03)}}
		, y label style = {at={(axis description cs:0.05,0.5)}}
		, xlabel={Number of instances}
		, ylabel={}
		, xmin=0, xmax=36
		, ymin=0, ymax=920
		, legend style={at={(0.74,0.98)},anchor=north, draw=none,fill=none}
		, legend columns=1
		, width=0.6\textwidth
		, height=0.5\textwidth
		, ytick={0,100,200,300,400,500,600,700,800,900}
		, xtick={0,6,12,18,24,30,36}
		, yticklabels={}
		, major tick length=2pt
		]
		
		\addplot [mark size=2.5pt, color=red, mark=diamond, dashed, mark options=solid] [unbounded coords=jump] table[col sep=semicolon, y index=7] {./cactus-ccp.csv}; 
		\addlegendentry{\vbs}
		
		\addplot [mark size=2.5pt, color=black, mark=*] [unbounded coords=jump] table[col sep=semicolon, y index=8] {./cactus-ccp.csv}; 
		\addlegendentry{\wasp \mccp}
		
		\addplot [mark size=2.5pt, color=black, mark=triangle*] [unbounded coords=jump] table[col sep=semicolon, y index=9] {./cactus-ccp.csv}; 
		\addlegendentry{\wasp \bccp}
		
		\addplot [mark size=2.5pt, color=blue, mark=diamond*] table[col sep=semicolon, y index=10] {./cactus-ccp.csv}; 
		\addlegendentry{\wasp \bfccp }
		
		\addplot [mark size=2pt, color=blue, mark=square*] table[col sep=semicolon, y index=11] {./cactus-ccp.csv};
		\addlegendentry{\wasp \bfaccp}
		
		\addplot [mark size=2.5pt, color=blue, mark=star] table[col sep=semicolon, y index=12] {./cactus-ccp.csv};
		\addlegendentry{\wasp \bfaccp (\pyth)}
		\end{axis}
		\end{tikzpicture}
	}
	
	\subfloat[]{\label{fig:eval:pup:scatter:enc1}%
		\begin{tikzpicture}[scale=0.7]
		\pgfkeys{%
			/pgf/number format/set thousands separator = {}}
		\begin{axis}[
		scale only axis
		, font=\small
		, x label style = {at={(axis description cs:0.5,0.03)}}
		, y label style = {at={(axis description cs:0.05,0.5)}}
		, xlabel={\vbs~ time (s)}
		, ylabel={\wasp~ \pred~ time (s)}
		, width=0.6\textwidth
		, height=0.5\textwidth
		, xmin=0, xmax=920
		, ymin=0, ymax=920
		, xtick={0,100,200,300,400,500,600,700,800,900}
		, ytick={0,100,200,300,400,500,600,700,800,900}
		, major tick length=2pt
		, title = {\enc{1} ({\color{blue}$\times$}) and \enc{2} ({\color{black}$+$})}
		]
		\addplot [mark size=4.5pt, only marks, color=blue, mark=x] [unbounded coords=jump] table[col sep=semicolon, x index=3, y index=4] {./scatter-pup.csv};
		\addplot [mark size=4.5pt, only marks, color=black, mark=+] [unbounded coords=jump] table[col sep=semicolon, x index=1, y index=2] {./scatter-pup.csv};
		\addplot [color=red, dashed] [unbounded coords=jump] table[col sep=semicolon, x index=0, y index=0] {./scatter-pup.csv};
		\end{axis}
		\end{tikzpicture}
	}
	\subfloat[]{\label{fig:eval:pup:barplot}%
		\begin{tikzpicture}[scale=0.7]
		\pgfkeys{%
			/pgf/number format/set thousands separator = {}}
		\begin{axis}[
		scale only axis
		, ybar	
		, font=\small
		, x label style = {at={(axis description cs:0.5,0.03)}}
		, y label style = {at={(axis description cs:0.05,0.5)}}
		, xlabel={\phantom{\vbs~ time (s)}}
		, width=0.6\textwidth
		, height=0.5\textwidth
		, xtick=data
		, major tick length=2pt
		, enlarge x limits=0.06
		, yticklabels={}
		, xticklabels={\textsc{cbc},\textsc{cplex},\textsc{cp},\textsc{sat},\vbs-\textsc{alt},\wasp \textsc{pred}}
		, nodes near coords={\pgfmathprintnumber\pgfplotspointmeta}
		, ylabel={}
		, title={\textsc{Solved instances}}
		]
		\addplot+[ybar] plot coordinates {(1,8) (2,10) (3,24) (4,25) (5,34) (6,36)};
		\end{axis}
		\end{tikzpicture}
	}
	\caption{Experimental analysis for \pup (a) and \ccp (b) instances. Instance by instance comparison for \pup using \enc{1} and \enc{2} encodings (c). Solved instances by approaches beyond ASP on \pup instances (d).} 
	\label{fig:eval}
	\figrule
\end{figure*}
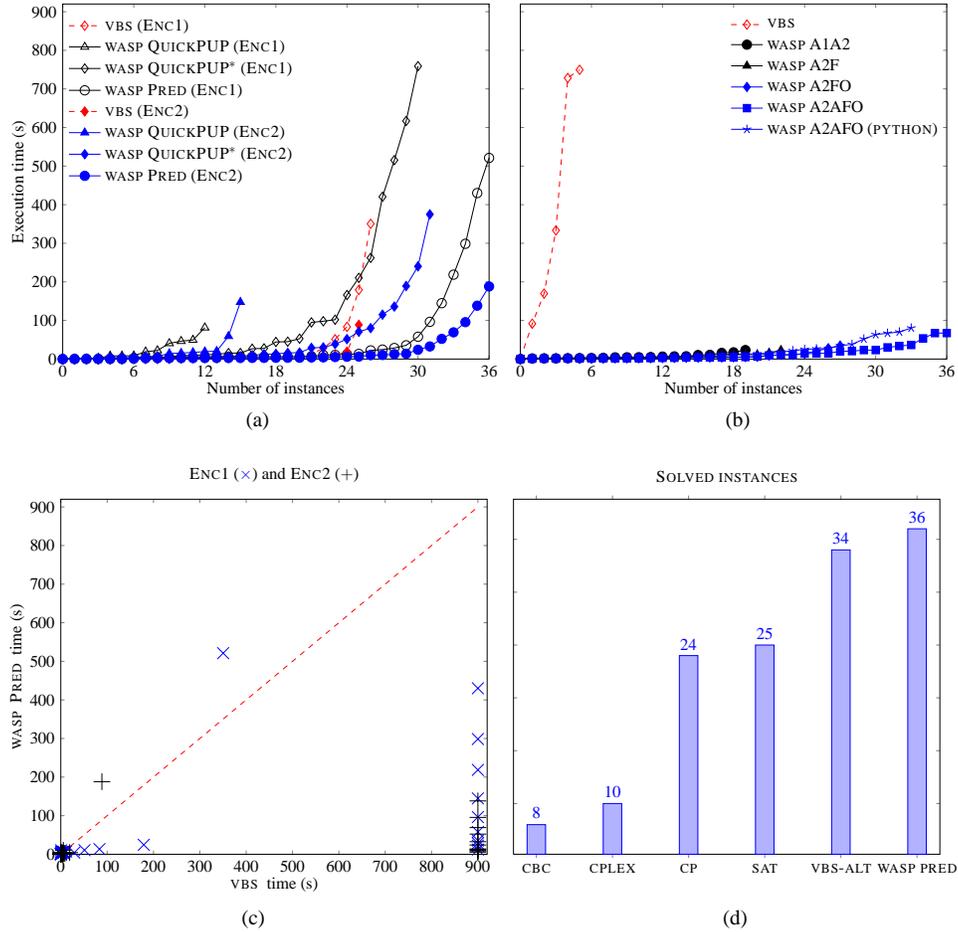

\myParagraph{Partner Units Problem.}
The evaluation results concerning \pup are summarized by means of a cactus plot in Figure~\ref{fig:eval:pup:cactus}. Recall that in a cactus plot a line is reported for each compared method, and there is a point $(x,y)$ in one of such lines whenever the corresponding method finds a solution for the $x^{th}$ fastest instance within $y$ seconds. 

\vbs was able to solve 26 and 25 instances using \enc{1} and \enc{2}, respectively.
\vbs failed for most of the \emph{double} as well as for some \emph{double-variant} and \emph{triple} instances.
Looking at Figure~\ref{fig:eval:pup:cactus}, note that \wasp with \qpup solved only 12 and 15 instances using \enc{1} and \enc{2} respectively.
The main reason for the discrepancy between our results and the ones obtained by~\cite{DBLP:conf/iaai/TeppanFF12} is due to the inability of the underlying ASP solver to generate new units on-the-fly. 
The number of available units is determined in the grounding step and cannot be changed later.
The original \qpup implementation, on the other hand, tries to quickly build a solution by creating and using new units whenever possible and reducing their number afterwards.
This strategy appears to be successful for greedy algorithms, but works poorly as a heuristic for \wasp. 
The remaining two heuristics, namely \qspup and \pred, are able to outperform \vbs with respect to the number of solved units.
Our best heuristic, \pred, solved \emph{all} instances independently of the encoding.
Furthermore, the solving time was reduced to at most 27 seconds using \enc{2}. 
An instance-wise comparison of \vbs with our best solution (i.e., \pred) is reported in the scatter plot in Figure~\ref{fig:eval:pup:scatter:enc1}, comparing results obtained with \enc{1} (points of shape ``$\times$'') and \enc{2} (points of shape ``$+$''). From the figure it is evident that \wasp with \pred outperforms \vbs in all instances but one, no matter the encoding used. 
In fact, the outlier instance (\texttt{2-triple-60.dl}) was solved in 188 and 88 seconds by \clasp with 10 threads using \enc{1} and \claspfolio with 10 threads using \enc{2}, whereas \pred needed 521 and 350 seconds, respectively.

We would like to emphasize the fact that \pred allowed \wasp to find solutions for all instances with the ``simple'' encoding \enc{1}. 
This is an important feature of our approach. 
It indicates that a programmer can use a non-optimized ASP modeling of a problem, and (if needed) scale the performance of the solver by embedding a domain-heuristic defined by a procedure. 

For the sake of completeness, we provide here a comparison of our best solution (\pred) with approaches alternative to ASP proposed in the literature.
In particular, we consider the same setting of \cite{DBLP:conf/cpaior/AschingerDFGJRT11}, where encodings for propositional satisfiability (\textsc{sat}), constraint programming (\textsc{cp}), and integer programming (\textsc{cplex} and \textsc{cbc}) were proposed. 
Results of the comparison are shown in Figure~\ref{fig:eval:pup:barplot}, where the number of solved instances is reported for each approach.
First of all, \textsc{sat} and \textsc{cp} outperform the approach based on integer programming, solving 25 and 24 instances, respectively. The virtual best combination of all the approaches alternative to ASP (\vbs-\textsc{alt}) allows to solve all instances but two within ten minutes.
This result is explained by the complementary behavior of \textsc{sat} and \textsc{cp}.
In fact, the former was effective on \emph{grid} instances but did not perform well on \emph{double}, and the other way round for \textsc{cp}.

\myParagraph{Combined Configuration Problem.}
The evaluation results concerning \ccp are summarized by means of a cactus plot in Figure~\ref{fig:eval:ccp:cactus}. 
Here, the reference system \vbs solved 5 out of 36 instances.
Instead, \wasp solved 22, 27 and all 36 instances with an average solving time of 6, 7, and 14 seconds using \bccp, \bfccp and \bfaccp, respectively. 
Moreover, since \bfaccp alternates between two heuristics, we performed an additional evaluation.
The goal was to test whether conflict learning is essential for finding a ``correct'' ordering as it was often the case for \pup heuristics.
Therefore, we saved the last ordering created by \bfaccp before a solution was found.
This ordering was used as initial one for another run of the solver.
In our experiments at least one more ordering, in addition to the saved one, was needed to find a solution.
Thus, we conclude that the interplay of the heuristic with the learning feature of CDCL is essential in that case.

\myParagraph{Scripting Interface Assessment.}
In Figure~\ref{fig:eval:ccp:cactus} we also report the results obtained by running a \pyth implementation of \bfaccp. It solves all instances but three within the timeout, and is on average slower than the performance-oriented \cpp implementation. 
The difference is due to several factors:
\begin{inparaenum}[(i)]
	\item the overhead of communication,
	\item a slower \pyth \textit{interpreter} (than compiled code), and
	\item some differences in the implementation and tuning of system libraries
\end{inparaenum}
Nonetheless, the \pyth version behaves acceptably and far better than \vbs.
Moreover, the \pyth version was \textit{considerably faster to develop than the \cpp one}. 
Indeed, according to the Function Point (FP) effort estimation metric~\cite{albrecht79:_fpa}, which is based on Lines of Code (LOC), \pyth and \perl require about 21 LOC/FP, whereas \cpp requires about 53 LOC/FP.
This is confirmed in our implementation where \cpp required about twice as much LOC compared to \pyth (1396 vs 631).

\section{Related Work}\label{sec:related}
Several ways of combining domain heuristics and ASP are proposed in the literature.

In \cite{DBLP:journals/aicom/Balduccini11}, a technique which allows learning of domain-specific heuristics in DPLL-based solvers is presented.
The basic idea is to analyze off-line the behavior of the solver on representative instances from the domain to learn and use a heuristic in later runs. 
In our approach, the heuristics are provided by domain-experts, nonetheless our extended \wasp could be considered as development platform for porting the ideas of \cite{DBLP:journals/aicom/Balduccini11} in a CDCL solver.

A declarative approach to definition of domain specific heuristics in ASP is presented in~\cite{DBLP:conf/aaai/GebserKROSW13}. 
The suggested \hclasp framework extends the \gringo language with a special \emph{\_heuristic} predicate. 
Atoms over this predicate allow one to influence the choices made by the default heuristic of \clasp. 
In fact, a user can provide initial weights, importance factors, decision levels and sign selection for atoms involved in non-deterministic decisions. 
\hclasp supports definition of dynamic heuristics by considering only those atoms over \emph{\_heuristic} predicate that are true in the current interpretation. 
In opposite to static heuristics, where heuristic decisions are encoded as facts, dynamic ones comprise normal rules with a \emph{\_heuristic} atom in the head.
As argued in~\cite{DBLP:conf/lpnmr/GebserRS15}, the grounding of programs comprising definitions of dynamic heuristics can be expensive as it impacts on grounding speed and size. 
The reason is that a grounder needs to output a rule for every possible heuristic decision.
Therefore, the authors avoid application of dynamic heuristics and use only a static one providing initial weights for a set of atoms.
Roughly, the idea of \cite{DBLP:conf/lpnmr/GebserRS15} is to compute a partial solution for an instance using a greedy algorithm. 
The obtained solution is then transformed to a set of heuristic atoms used to initialize the counters of the default heuristics. Thus, \clasp tries the partial solution first. 
A similar behavior could be obtained in our implementation by properly defining the functions \I, \F, \S of the command \response{Fallback}{$0, \I, \F, \S$} (see Table \ref{tab:eventscommands}).

The main drawback of this approach is that the applied heuristic is static and cannot be updated during the computation.
On the contrary, our variants of the same heuristics (e.g., compare \mccp with \bfaccp in Section~\ref{sec:ccpheur}) have been designed to be dynamic, i.e. they are queried multiple times during the execution of the algorithm so to consider different partial solutions to start with.
The source code of~\cite{DBLP:conf/lpnmr/GebserRS15} is unavailable, thus a direct comparison with their implementation via re-evaluation was not possible.
Nonetheless, since their empirical evaluation was done on a similar hardware and on a subset of the instances we considered, we report that in \cite{DBLP:conf/lpnmr/GebserRS15} the implementation was able to solve 6 out of the 20 \textit{real-world instances} (see Section~\ref{sec:experiments}) whereas \wasp with \bfaccp solved all of them. 

Additionally, we also evaluated the performance of \hclasp on dynamic heuristics. Therefore, we implemented an ASP program for the best-fit heuristic~\cite{DBLP:books/fm/GareyJ79} of the bin packing problem used in \accp{2}.
Evaluation on the CCP instances presented in Section~\ref{sec:experiments} showed impracticality of this approach and confirmed similar observation of ~\cite{DBLP:conf/lpnmr/GebserRS15}.
For the sake of completeness, we report that the grounding step alone requires up to 4 minutes (1 minute on average), but resulted in ground programs up to 2 Gb (300 Mb on average).
The vast majority of ground rules in these programs represent all possible heuristic decisions for any partial interpretation. 
Since such tabular representation of the best-fit part of the \accp{2} heuristic alone is already very large, we discontinued this experiment. 
Next, we researched on possible improvements due to \clasp \texttt{--dom-mod} option, which also introduces domain modifications to the default heuristic. 
The best result in our test -- solving one instance -- was obtained using the value \texttt{(3,16)}, which forces the internal \clasp heuristic to assign \emph{true} to all atoms specified with the \texttt{\#show} directive of the \gringo language.

\pup was introduced in \cite{DBLP:journals/aiedam/FalknerHSS08} where a number of different approaches were considered: Alloy, ASP, CSP, 
Constraint Handling Rules as well as problem-specific algorithm based on graph partitioning. 
\cite{DBLP:journals/aiedam/FalknerHSS08} reports that all these approaches are not effective on large industrial instances.
A thorough modeling and evaluation of \pup was continued in \cite{DBLP:conf/cpaior/AschingerDFGJRT11}. 
The results reported in Section~\ref{sec:experiments} confirm the findings of \cite{DBLP:conf/cpaior/AschingerDFGJRT11}: None of the applied general methods are able to solve all the instances. This result pushed the research towards investigation of domain-specific heuristics.
These efforts resulted in development of the \qpup heuristic \cite{DBLP:conf/iaai/TeppanFF12}. 
The suggested greedy algorithm is based on backtrack search combined with \qpup for variable orderings and time-based restarts.
Later, results of \cite{DBLP:conf/ictai/Drescher12} witnessed that the performance of declarative approaches can be also improved using similar heuristics. 
The author suggested \pred heuristic and implemented it using both $\text{ECL}^i\text{PS}^e$ Prolog and the CSP solver \textsc{Minion}. 
The first approach included a custom propagator for the partner unit constraint and a Prolog definition of the heuristic. The second one utilized a very efficient implementation of a \textit{global cardinality constraint} as it was unfeasible to implement \pred in \textsc{Minion}. 
Therefore, the solvers were extended with specific implementations of procedural propagation or decision-making techniques. 
The source code of \cite{DBLP:conf/ictai/Drescher12} is not available, thus also in this case a re-evaluation was not possible. 
However, \cite{DBLP:conf/ictai/Drescher12} report that some of the triple and grid with \iucap=4
were not solved, whereas our approach solved all of them. 

\section{Conclusion}\label{sec:conclusion}
In this paper the combination of ASP and domain-specific heuristics is studied for solving two challenging industrial applications of ASP: the Partner Units and the Combined Configuration problems.
Several known heuristic criteria and also novel ones are considered and implemented on the top of the ASP solver \wasp. 
As a by-product of this work, \wasp was extended with an interface that eases the development of new external heuristics. 
The interface was effective and simplified the development of the heuristics presented in this work. The benefits  in terms of development time (w.r.t. the first ad-hoc modifications) were sensible; for instance the \pyth version of heuristic \bfaccp has been implemented in only one working day.
Moreover, the performances of our implementations have been evaluated on all the real-world instances of \pup and \ccp ever provided by Siemens.
Results are positive: \wasp solves all the considered instances within few minutes outperforming all known alternatives. 
As future and ongoing work we are investigating the application of ASP and domain-heuristics for solving a hard resource assignment problem modeling a requirement of a Siemens project, for solving the allotment problem~\cite{DBLP:conf/rr/DodaroLNR15}, and for performing complex reasoning on combinatorial auctions \cite{DBLP:journals/ai/FiondaG13}.

\section{Acknowledgement}
This work was partially supported by the Austrian Science Fund (FWF) contract number I 2144 N-15, the Carinthian Science Fund (KWF) contract KWF-3520/26767/38701, the Italian Ministry of University, Research under PON project ``Ba2Know (Business Analytics to Know) Service Innovation -- LAB'', No. \ PON03PE\_00001\_1, and by the Italian Ministry of Economic Development under project ``PIUCultura (Paradigmi Innovativi per l'Utilizzo della Cultura)'' n.\ F/020016/01--02/X27.

\bibliographystyle{acmtrans}

\clearpage
\appendix
\section{Heuristics Development Example}
\label{appendix:pigeon}
\begin{figure}[t!]
\begin{lstlisting}[mathescape=true]
# Input:
# * a set of pigeons P={1,...,n}, defined by means of the predicate pigeon
# * a set of holes$\phantom{lll}$ H={1,...,m}, defined by means of the predicate hole
pigeon(1) $\leftarrow$  $\phantom{addspace}$ hole(1) $\leftarrow$
$\phantom{add}\vdots$ $\phantom{addlooooooootofspace}$ $\vdots$
pigeon(n) $\leftarrow$ $\phantom{addspace}$ hole(m) $\leftarrow$

# Guess an assignment
inHole($p$,$h$) $\leftarrow$ not outHole($p$,$h$) $\forall p \in P,\, \forall h \in H$
outHole($p$,$h$) $\leftarrow$ not inHole($p$,$h$) $\forall p \in P,\, \forall h \in H$

# A hole contains at most one pigeon
$\leftarrow$ inHole($p_i$,$h$), inHole($p_j$,$h$) $\forall p_i,p_j \in P \mid i \neq j,\, \forall h \in H$

# A pigeon is assigned to at most one hole
$\leftarrow$ inHole($p$,$h_i$), inHole($p$,$h_j$) $\forall h_i,h_j \in H \mid i \neq j,\, \forall p \in P$

# A pigeon must be in some hole
inSomeHole($p$) $\leftarrow$ inHole($p$,$h$) $\forall p \in P,\, \forall h \in H$
$\leftarrow$ not inSomeHole($p$) $\forall p \in P$
\end{lstlisting}
\caption{ASP encoding of Pigeonhole.}
\label{fig:pigeonasp}
\vspace{1em}
		
\begin{lstlisting}[language=Python]
# global data structures
var = {1: 'false', 'false': 1}
P = []  # list of all pigeon-constants
H = []  # list of all hole-constants

def addedVarName(v, name):
    #invoked when WASP parses the atom table of the gringo numeric format
    global var, H, P
    var.update({v: name, name: v})
    if name.startswith("pigeon"):
        P.append(name[7:-1])
    if name.startswith("hole"):
        H.append(name[5:-1])

def onFinishedParsing():
    # disable simplifications of variables
    return [v for v in var.keys() if isinstance(v, int)]

def choiceVars(): #event onChoiceRequired, invoked when a choice is needed
    global var, H, P
    if len(P) > len(H):
        return [4, 0]  # force incoherence
    # assign pigeon i to hole i
    return [var["inHole(%s,%s)" % (i, i)] for i in range(1, len(P))]

def onChoiceContradictory(choice): #event onIncoChoice
    pass  # no choice can be contradictory
\end{lstlisting}
\caption{Pigeonhole heuristic in \pyth.}
\label{fig:python}
\end{figure}

In order to exemplify the usage of the infrastructure we report here the solution of the well-known Pigeonhole problem, whose ASP encoding is reported in Figure~\ref{fig:pigeonasp}.
Albeit for this toy problem the solution is trivial, modern ASP solvers fail to recognize efficiently when an instance admits no solutions.
It is easy to see that when the number of pigeons exceeds the number of holes, no solution can be found.
Otherwise, a solution can be easily obtained by associating the $i$-$th$ pigeon with the $i$-$th$ hole.

A heuristic strategy based on this observation can be implemented using \pyth as reported in Figure~\ref{fig:python}.
First, we initialize the global data structures \verb|var|, \verb|H|, and \verb|P| for storing the association of atom names to a numeric identifier created by \gringo, the set of holes, and the set of pigeons, respectively.
The method \verb|addedVarName| is called whenever a new variable \verb|v| named \verb|name| is added inside \wasp.
Here, we store the association between the variable identifier \verb|v| and its name \verb|name| and vice-versa.
Moreover, we check whether the variable represents a pigeon or a hole by checking the name.
If this is the case the ASP constant representing the pigeon or hole is added \verb|P| or \verb|H|, respectively.

After the parsing of the input program, the method \verb|onFinishedParsing| is invoked. This method is allowed to return a list of variable that must be \emph{frozen}, i.e. variables that must not be removed during the simplification step.
In our example, all variables are frozen.

Later on, \wasp searches for an answer set.
During the computation, the method \verb|choiceVars| is invoked whenever a choice is needed (\event{ChoiceRequired}{~}).
This method may return:
\begin{itemize}
	\item A literal representing the next choice (command \response{Choose}{$\ell$}).
	\item A list of literals representing the next choices (command \response{Choose}{$\ell$} repeated for all literals in the list).
	\item Special values representing other commands. In particular, \verb|[4, 0]| is used to stop the computation returning \inco (command \response{AddConstraint}{$\leftarrow \naf \bot$}).
\end{itemize}
In our example, the method \verb|choiceVars| first checks whether the holes are sufficient to host all pigeons.
If this is not the case it returns \verb|[4,0]|. Otherwise, it returns a list of choices where the atoms \verb|inHole(i,i)| ($i \in [1, \ldots, |P|]$) will be set to true.

Finally, the method \verb|onChoiceContradictory| (event \event{IncoChoice}{$\ell$}) is invoked whenever a previous choice lead to an inconsistency.
In our case, this method performs no operation since none of the choices can be contradictory.

\section{Additional Plots}
\label{appendix:detailedplots}
In this section, we present additional cactus plots comparing all off-the-shelf ASP solvers considered in our experiments with our best heuristic variant for each problem.
In particular, Figures~\ref{fig:appendix:pup:orig} and \ref{fig:appendix:pup:m2} report on the performance of the solvers on PUP instances employing encoding \enc{1} and \enc{2}, respectively.
Concerning \enc{1}, we observe that \clasp with 10 threads and \claspfolio obtained similar performance solving 25 instances. All other systems solve 22 instances.
Similar considerations hold also for \enc{2}, where the best versions are \clasp with 10 threads and \clasp with portfolio also solve 25 instances each.
Looking at the \vbs lines, we observe that all solvers behave similarly solving basically the same set of instances. However, in case of \enc{1}, \vbs was able to solve one instance more than using \enc{2} (\texttt{2-doublev-120.dl}). This is due to optimizations, like symmetry breaking rules, towards the grid and triple-like instances included in \enc{2}. On one hand, \vbs with \enc{2} required far less time to find solutions for grids and triples. On the other hand, due to these optimizations the performance on instances of the double and double-variant types was slightly lower than using \enc{1}. As a result, \clasp portfolio was able to solve one instance more, thus improving the overall performance of \vbs. More detailed results can be found on the companion website~\url{http://yarrick13.github.io/hwasp/}.

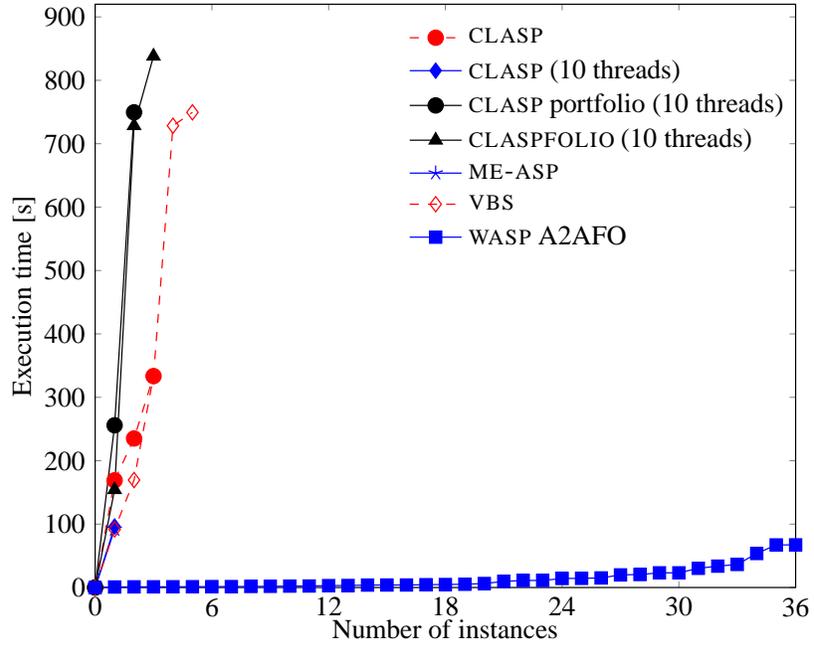
\begin{figure}[p]
	\figrule
  	\begin{tikzpicture}[scale=1.15]
  	\pgfkeys{%
  		/pgf/number format/set thousands separator = {}}
  	\begin{axis}[
  	scale only axis
  	, font=\small
  	, x label style = {at={(axis description cs:0.5,0.04)}}
  	, y label style = {at={(axis description cs:0.05,0.5)}}
  	, xlabel={Number of instances}
  	, ylabel={Execution time [s]}
  	, xmin=0, xmax=36
  	, ymin=0, ymax=920
  	, legend style={at={(0.72,0.98)},anchor=north, draw=none,fill=none}
  	, legend columns=1
  	, width=0.6\textwidth
  	, height=0.5\textwidth
  	, ytick={0,100,200,300,400,500,600,700,800,900}
  	, xtick={0,6,12,18,24,30,36}
  	, major tick length=2pt
  	]
  	
  	\addplot [mark size=2.5pt, color=red, mark=*, dashed, mark options=solid] [unbounded coords=jump] table[col sep=semicolon, y index=1] {./plot-data-ccp-appendix.csv}; 
  	\addlegendentry{\clasp}
  	
  	\addplot [mark size=2.5pt, color=blue, mark=diamond*] table[col sep=semicolon, y index=4] {./plot-data-ccp-appendix.csv}; 
  	\addlegendentry{\clasp (10 threads)}
  	
  	\addplot [mark size=2.5pt, color=black, mark=*] [unbounded coords=jump] table[col sep=semicolon, y index=2] {./plot-data-ccp-appendix.csv}; 
  	\addlegendentry{\clasp portfolio (10 threads)}
  	
  	\addplot [mark size=2.5pt, color=black, mark=triangle*] [unbounded coords=jump] table[col sep=semicolon, y index=3] {./plot-data-ccp-appendix.csv}; 
  	\addlegendentry{\claspfolio (10 threads)}
  	
  	\addplot [mark size=2.5pt, color=blue, mark=star] table[col sep=semicolon, y index=5] {./plot-data-ccp-appendix.csv};
  	\addlegendentry{\measp}
  	
  	\addplot [mark size=2.5pt, color=red, mark=diamond, dashed, mark options=solid] [unbounded coords=jump] table[col sep=semicolon, y index=7] {./cactus-ccp.csv}; 
  	\addlegendentry{\vbs}
  	
  	\addplot [mark size=2pt, color=blue, mark=square*] table[col sep=semicolon, y index=6] {./plot-data-ccp-appendix.csv};
  	\addlegendentry{\wasp \bfaccp}
  	
  	\end{axis}
  	\end{tikzpicture}
      \caption{Comparison of all solvers on CCP instances \label{fig:appendix:ccp}}
  \figrule
\end{figure}

Concerning CCP, the performance of the solvers is reported in Figure~\ref{fig:appendix:ccp}.
Here, all considered solvers solve at most 5 instances.

As a general comment we observe that multi-threaded versions exploiting 10 times more hardware resources and several heuristics are slightly better than single-threaded alternatives. Nonetheless, our heuristic variants are faster than all other alternatives.

\begin{figure}[b]
  \figrule
    \begin{tikzpicture}[scale=1.15]
    \pgfkeys{%
    	/pgf/number format/set thousands separator = {}}
    \begin{axis}[
    scale only axis
    , font=\small
    , x label style = {at={(axis description cs:0.5,0.04)}}
    , y label style = {at={(axis description cs:0.05,0.5)}}
    , xlabel={Number of instances}
    , ylabel={Execution time [s]}
    , xmin=0, xmax=36
    , ymin=0, ymax=920
    , legend style={at={(0.30,0.98)},anchor=north, draw=none,fill=none}
    , legend columns=1
    , width=0.6\textwidth
    , height=0.5\textwidth
    , ytick={0,100,200,300,400,500,600,700,800,900}
    , xtick={0,6,12,18,24,30,36}
    , major tick length=2pt
    ]
    
    \addplot [mark size=2.5pt, color=red, mark=*, dashed, mark options=solid] [unbounded coords=jump] table[col sep=semicolon, y index=1] {./plot-data-pup-org-appendix.csv}; 
    \addlegendentry{\clasp}
    
    \addplot [mark size=2.5pt, color=blue, mark=diamond*] table[col sep=semicolon, y index=4] {./plot-data-pup-org-appendix.csv}; 
    \addlegendentry{\clasp (10 threads)}
    
    \addplot [mark size=2.5pt, color=black, mark=*] [unbounded coords=jump] table[col sep=semicolon, y index=2] {./plot-data-pup-org-appendix.csv}; 
    \addlegendentry{\clasp portfolio (10 threads)}
    
    \addplot [mark size=2.5pt, color=black, mark=triangle*] [unbounded coords=jump] table[col sep=semicolon, y index=3] {./plot-data-pup-org-appendix.csv}; 
    \addlegendentry{\claspfolio (10 threads)}
    
    \addplot [mark size=2.5pt, color=blue, mark=star] table[col sep=semicolon, y index=5] {./plot-data-pup-org-appendix.csv};
    \addlegendentry{\measp}
    
    \addplot [mark size=2.5pt, color=red, mark=diamond, dashed, mark options=solid]  table[col sep=semicolon, y index=16] {./cactus-pup.csv}; 
    \addlegendentry{\vbs}
    
    \addplot [mark size=2pt, color=blue, mark=square*] table[col sep=semicolon, y index=6] {./plot-data-pup-org-appendix.csv};
    \addlegendentry{\wasp \pred}
    
    \end{axis}
    \end{tikzpicture}
      \caption{Comparison of all solvers on PUP instances (\enc{1})\label{fig:appendix:pup:orig}}
    \figrule
\end{figure}

\begin{figure}[p]
	\figrule
	\begin{tikzpicture}[scale=1.15]
	\pgfkeys{%
		/pgf/number format/set thousands separator = {}}
	\begin{axis}[
	scale only axis
	, font=\small
	, x label style = {at={(axis description cs:0.5,0.04)}}
	, y label style = {at={(axis description cs:0.05,0.5)}}
	, xlabel={Number of instances}
	, ylabel={Execution time [s]}
	, xmin=0, xmax=36
	, ymin=0, ymax=920
	, legend style={at={(0.30,0.98)},anchor=north, draw=none,fill=none}
	, legend columns=1
	, width=0.6\textwidth
	, height=0.5\textwidth
	, ytick={0,100,200,300,400,500,600,700,800,900}
	, xtick={0,6,12,18,24,30,36}
	, major tick length=2pt
	]
	
	\addplot [mark size=2.5pt, color=red, mark=*, dashed, mark options=solid] [unbounded coords=jump] table[col sep=semicolon, y index=1] {./plot-data-pup-m2-appendix.csv}; 
	\addlegendentry{\clasp}
	
	\addplot [mark size=2.5pt, color=blue, mark=diamond*] table[col sep=semicolon, y index=4] {./plot-data-pup-m2-appendix.csv}; 
	\addlegendentry{\clasp (10 threads)}
	
	\addplot [mark size=2.5pt, color=black, mark=*] [unbounded coords=jump] table[col sep=semicolon, y index=2] {./plot-data-pup-m2-appendix.csv}; 
	\addlegendentry{\clasp portfolio (10 threads)}
	
	\addplot [mark size=2.5pt, color=black, mark=triangle*] [unbounded coords=jump] table[col sep=semicolon, y index=3] {./plot-data-pup-m2-appendix.csv}; 
	\addlegendentry{\claspfolio (10 threads)}
	
	\addplot [mark size=2.5pt, color=blue, mark=star] table[col sep=semicolon, y index=5] {./plot-data-pup-m2-appendix.csv};
	\addlegendentry{\measp}
	
	\addplot [mark size=2.5pt, color=red, mark=diamond, dashed, mark options=solid]  table[col sep=semicolon, y index=6] {./cactus-pup.csv};
	\addlegendentry{\vbs}
	
	\addplot [mark size=2pt, color=blue, mark=square*] table[col sep=semicolon, y index=6] {./plot-data-pup-m2-appendix.csv};
	\addlegendentry{\wasp \pred}
	\end{axis}
	\end{tikzpicture}
	
      \caption{Comparison of all solvers on PUP instances (\enc{2})\label{fig:appendix:pup:m2}}
	\figrule
\end{figure}

\end{document}